\documentclass[conference]{IEEEtran}

\newif\ifreview

\IEEEoverridecommandlockouts
\usepackage{cite}
\usepackage{amsmath,amssymb,amsfonts}
\usepackage{algorithmic}
\usepackage{graphicx}
\usepackage{textcomp}
\usepackage{xcolor}
\def\BibTeX{{\rm B\kern-.05em{\sc i\kern-.025em b}\kern-.08em
    T\kern-.1667em\lower.7ex\hbox{E}\kern-.125emX}}

\usepackage{diagbox}
\usepackage{pifont}
\usepackage{makecell}
\usepackage{array}   
\usepackage{multirow}
\usepackage{booktabs}

\usepackage{hyperref} 

\DeclareMathOperator*{\argmax}{argmax}
    
\begin{document}

\title{How to Recognize New Words: A Comparison Between Context Biasing Methods and Speech LLMs
}

\ifreview
\author{
  \IEEEauthorblockN{Anonymous Authors}
  \IEEEauthorblockA{
    \textit{Anonymous Institution} \\
    City, Country \\
    anonymous@institution.example
  }
}
\else
\author{\IEEEauthorblockN{1\textsuperscript{st} Christian Huber}
\IEEEauthorblockA{\textit{Interactive Systems Lab} \\
\textit{Karlsruhe Institute of Technology}\\
Karlsruhe, Germany\\
christian.huber@kit.edu}
\and
\IEEEauthorblockN{2\textsuperscript{nd} Alexander Waibel}
\IEEEauthorblockA{\textit{Interactive Systems Lab} \\
\textit{Carnegie Mellon University}\\
Pittsburgh PA, USA\\
alexander.waibel@cmu.edu}
}
\fi

\maketitle

\begin{abstract}
Recognizing new and rare words - named entities, acronyms, domain specific special words, and other items scarce in training data - remains a key challenge for automatic speech recognition (ASR). We compare two strategies for this: context biasing methods, where an ASR model is extended such that during inference a word list can be supplied, and speech large language models (LLMs) prompted with context directly. We evaluate two context biasing methods based on Whisper against three speech LLMs across read and non-read speech, reporting biased, unbiased, and overall word error rate (WER). The context biasing methods cut biased WER by up to 88\% relative while leaving other words largely unaffected. Speech LLMs excel on read speech but generalize less well to non-read speech, and prove sensitive to distractor count and prompt word order. We characterize the resulting trade-offs to guide method selection.
\end{abstract}

\begin{IEEEkeywords}
context biasing, speech LLMs, new words
\end{IEEEkeywords}

\section{Introduction}

End-to-end ASR has advanced rapidly, driven by the transformer architecture \cite{vaswani2017attention,pham2019very} and, more recently, by large models trained with weak supervision on hundreds of thousands of hours of audio, such as Whisper\cite{radford2023robust}. These systems achieve low word error rates across a wide range of conditions and have become the default choice for many transcription tasks. Nevertheless, they share a persistent weakness: words that are rare or absent in the training data - named entities, acronyms, and domain-specific terminology - are recognized far less reliably than common words.
This weakness is consequential because rare words are often the most informative parts of an utterance. A person's name, a product identifier, or a technical term typically carries more meaning than the surrounding function words, so an error on such a word degrades the usefulness of a transcript disproportionately. At the same time, because these words are infrequent, their errors are diluted in the overall WER, which can stay low even when the words that matter most are systematically missed.

Two strategies can be used to address this. The first, context biasing methods, which extends an ASR model with a mechanism that accepts a list of words or phrases at inference and biases decoding toward them, without altering the model's parameters for each new list. The second exploits speech LLMs, a recent class of models that jointly consume audio and a text prompt: the words of interest - or even a free-form description of the context - can be placed in the prompt, and the model can attend to them. Context biasing methods offers a dedicated, controllable interface, whereas speech LLMs promise greater flexibility. 


\section{Related Work}

The difficulty of recognizing new and rare words is long-standing.
Some works have combined statistical or neural language models with end-to-end ASR models using shallow fusion \cite{sriram2017cold, williams2018contextual, kannan2018analysis, huang2020class, kojima2022study}.
Because the biasing component is trained separately from the acoustic model, the gains are typically modest.
On the other hand, many recent works have used attention-based deep biasing
\cite{pundak2018deep, bruguier2019phoebe, jain2020contextual, huber2021instant, le2021contextualized, han2022improving, dingliwal2023personalization, huang2023contextualized, yang2024promptasr, sudo2024contextualized, yu2024lcb, xiao2025contextual,sudo2025owsm}. Some of them use only textual context information and some also include pronunciation information.

Recently, speech LLMs were introduced\cite{chu2023qwen,tang2024salmonn,hu2024wavllm}. The audio signal is encoded and the resulting representation is fed in an LLM. Some works adapt a speech LLM to include context\cite{lakomkin2024end, gong2024contextual
}. The latest speech LLMs\cite{Qwen3-ASR,xu2025qwen3,peng2026vibevoice} are explicitly trained to use the context given in the prompt.
Closest to our work is \cite{sudo2025owsm}, however, they do not study the effect of distractor count for large speech LLMs and only evaluate on read speech (LibriSpeech).

In this paper, we (1) present a controlled, head-to-head comparison of two Whisper-based context biasing methods against three recent speech LLMs (Qwen3-ASR, Qwen3-Omni, VibeVoice-ASR) on rare-word recognition and report with the biased, unbiased, and overall WER metrics, (2) evaluate across read (LibriSpeech) and non-read (Earnings-21, Yodas) speech, showing that speech LLMs are strong on read speech but generalize less well to non-read speech, (3) systematically probe robustness, quantifying sensitivity to distractor count and sensitivity to the order of words in the prompt, (4) study a filtering stage that uses an auxiliary speech LLM to prune the bias list, showing how much of the loss it recovers. (5) From these results we characterize the resulting trade-offs to guide method selection: dedicated context biasing methods remain the more robust choice when a clean word list is available, whereas speech LLMs offer a more flexible interface at the cost of prompt sensitivity and an extra filtering stage.

\section{Background}
\label{sec:background}

In the following paragraphs we summarize the two context biasing methods we compare.

An auto-regressive end-to-end ASR model directly estimates the probability distribution
\begin{equation}
    P(Y_t|Y_0,\ldots,Y_{t-1};X)
    \label{eq:1}
\end{equation}
of the next token $Y_t$ given the already decoded sequence $Y_0,\ldots,Y_{t-1}$ and the audio input $X$.
The model is then used to find the word sequence $\hat{Y}$ with the highest probability
\begin{align*}
    \hat{Y}&=\argmax_{Y}P(Y|X)\\
    &=\argmax_{Y}\prod_{t=1}^TP(Y_t|Y_0,\ldots,Y_{t-1};X),
\end{align*}
where $Y_0$ is the start of sequence token.

We work with a transformer-based encoder-decoder ASR model. First, $Y_0,\ldots,Y_{t-1}$ is embedded:
\begin{equation}
    E=Emb(Y_0,\ldots,Y_{t-1})\in\mathbb{R}^{t\times d}, \label{eq:emb}
\end{equation}
with $d\in\mathbb{N}$. Then, the decoder output is computed:
\begin{equation*}
O=Dec(H_X,E)\in\mathbb{R}^{t\times d},
\end{equation*}
where $H_X=Enc_{Audio}(X)$ is the encoded audio input. Finally, the output and softmax layers are applied:
\begin{equation*}
    \alpha=Linear(O)\in\mathbb{R}^{t\times n_{vocab}}, \label{eq:logits}
\end{equation*}
\begin{equation}
    p=Softmax(\alpha)\in\mathbb{R}^{t\times n_{vocab}}, \label{eq:lprobs}
\end{equation}
where $n_{vocab}$ is the vocabulary size.

Based on that model, we trained two context biasing models using the training scheme from \cite{huber2021instant} together with the architecture from either \cite{huber2021instant} (denoted A) or \cite{sudo2025owsm} (denoted B).
The methods works as follows:
Equation \ref{eq:1} is replaced by
\begin{equation*}
    P(Y_t|Y_0,\ldots,Y_{t-1};X;Z),
\end{equation*}
where $Z$ is some context provided to the model. In our case,
\begin{equation*}
Z=(Z_1,\ldots,Z_L), L\in\mathbb{N},
\end{equation*}
is a list denoted context biasing list and each $Z_l$, $l\in\{1,\ldots,L\}$, is a word or short phrase the model is biased towards.

\subsection{Context encoding}


Both models incorporate the context biasing list by first
encoding and embedding each item. 
The result is denoted
\begin{equation*}
    Z_l^{emb} = Emb(Tokenize(Z_l)).
\end{equation*}
Then,
an 
encoder is applied independently for each item, and afterward the mean over the sequence dimension is computed.
This results in one vector per list entry, denoted
\begin{equation*}
    Z_l^s = Avg(Enc(Z_l^{emb})),
\end{equation*}
which can be interpreted as a summary vector of the list entry.
For method A, a learned dummy vector $Z_0^s$ is added, which is later used to determine when there is no relevant information in the context biasing list.

\subsection{Context decoding}

\textbf{Method A:}
Each decoder layer of the baseline model consists of a self-attention layer, a cross-attention layer and a positionwise feed-forward layer.
After each cross-attention layer, a context-attention layer is added, which as first step predicts the availability/location of relevant information in the context biasing list (see next paragraph on how to train that task).
If there is relevant information, it is extracted in the second step using attention (see equation \ref{eq:2}):
Let $I^m$,
$m\in\{1,\ldots,M\}$,
$M\in\mathbb{N}$,
be the input of the $m$-th context-attention layer. 
First, $I^m$ is used as a query and $(Z_l^s)_{l=0,\ldots,L}$ as keys to compute similarity scores $S^m=I^m\cdot(Z_l^s)_{l=0,\ldots,L}^T$.
Then, the output $O^m$ of the $m$-th context-attention layer is computed as follows, where $n_t=\argmax_{0\le l\le L} S_{t,l}^m$:
\begin{equation}
    O_t^m=I_t^m+\begin{cases}Attn(q=I_t^m, k=Z_{n_t}^{emb}, v=Z_{n_t}^{emb}), &{n_t}>0,\\
    0, &{n_t}=0.
    \end{cases}
    \label{eq:2}
\end{equation}

Using only the most relevant context biasing list item (because of the argmax) for each token has the advantage that the result is independent from
items which are irrelevant.

\textbf{Method B:}
For this method $Z^s$ is used to extend the vocabulary of the decoder. This is done by extending the output layer, which maps the output of the final decoder layer to the vocabulary, and extending the embedding layer.

In particular,
\begin{equation}
    \alpha_{Context}=\frac{Linear_2(O)\cdot Linear_3(Z^s)^T}{\sqrt d}\in\mathbb{R}^{t\times L}. \label{eq:acontext}
\end{equation}
is calculated and $\alpha$ in equation \ref{eq:lprobs} is replaced by
\begin{equation*}
    Concat(\alpha,\alpha_{Context})\in\mathbb{R}^{t\times (n_{vocab}+L)}.
\end{equation*}
Furthermore, $Y_0,\ldots,Y_{t-1}$ is replaced by $Y'_0,\ldots,Y'_{t-1}$, where $Y'_0,\ldots,Y'_{t-1}$ is calculated by replacing all subsequences of $Y_0,\ldots,Y_{t-1}$ which correspond to a context biasing list entry $Z_l$ with a dynamic token $v_l$.
Finally, $E$ in equation \ref{eq:emb} is replaced by
\begin{equation*}
E'=Emb(Y'_0,\ldots,Y'_{t-1}),
\end{equation*}
where dynamic tokens $v_l$ are embedded by $Linear_4(Z_l^s)$ and the rest of the tokens is embedded using $Emb$.

\subsection{Training}

During model training, in each step, the context bias list $Z$ is sampled from the labels of the corresponding batch. 
 
Therefore, for each token $Y_t$ the ground truth item $G_t\in\{0,\ldots,L\}$ in the context biasing list is known, where $G_t=0$ means that for that token no relevant information is available in the context biasing list. 
For model A, the $loss$ consists of two cross-entropy parts.
The first term classifies the next token $Y_t$
of the sequence
and the second term guides the network towards relevant context biasing list items:
\begin{align*}
    loss = &\ \frac{1}{T}\sum_{t=1}^TCE(f_\theta(Y_0,\ldots,Y_{t-1};X;Z),Y_t)\\
    &+ \frac{\lambda}{MT} \sum_{m=1}^M \sum_{t=1}^TCE(S_t^m,G_t),
\end{align*}
\noindent where $f_\theta(Y_0,\ldots,Y_{t-1};X;Z)$ is the network output,
$CE$ the cross-entropy loss function and $\lambda>0$ a hyperparameter.

For model B, only the first $loss$ term is present.

Specifically, we used a batch size of 16 and sampled on average three context biasing list entries per utterance of the batch. Then the context biasing list is filled up to a length of 200 with distractors sampled from other batches.

\section{Experiments}

\subsection{Models}
\label{sec:models}

We compare the two context biasing methods explained in Section \ref{sec:background} with three speech LLMs: Qwen3-ASR \cite{Qwen3-ASR} (1.7B, denoted C), Qwen3-Omni \cite{xu2025qwen3} (30B, denoted D) and VibeVoice-ASR \cite{peng2026vibevoice} (9B, denoted E).

We use Whisper \cite{radford2023robust} (whisper-large-v2) as the baseline ASR model for both context biasing methods. The context biasing list is tokenized / embedded using the Whisper tokenizer / Whisper embedding, and
the context is encoded using the mBART-50 encoder \cite{tang2020multilingual}.
We trained both context biasing models on Common voice \cite{ardila2019common} and only the newly added parameters of the mBART-50 encoder and the context-attention layers (if present) were adapted.


Qwen3-ASR was trained in the supervised finetuning stage with "context biasing data. [...] the model learns to utilize the context tokens inside the system prompt as background knowledge, allowing users to obtain customized ASR results"\cite{Qwen3-ASR}.
Qwen3-Omni is promoted as "a single multimodal model that for the first time maintains state-of-the-art performance across text, image, audio, and video without any degradation relative to single-modal counterparts"\cite{xu2025qwen3}.
When using VibeVoice-ASR, users can "supply customized context - ranging from hotword lists to background descriptions - significantly enhancing the model’s ability to recognize domain-specific terminology"\cite{peng2026vibevoice}.

During decoding of the test sets (see Section \ref{sec:data}), the context biasing list / prompt contains the rare words belonging to the utterance which is currently decoded. Optionally, we add up to 250 distractors chosen randomly from the other rare words of the testset and / or remove the relevant context.
The context biasing methods are order-agnostic with respect to the context biasing list. This is not the case for the speech LLMs. We found that they are sensitive with respect to the order in which the words from the context biasing list are placed in the prompt. In particular, the speech LLMs perform better if a word is at the beginning of the prompt. To ensure a fair comparison, we sort the words in the prompt alphabetically for all speech LLMs.

\begin{table*}

\caption{Results: BWER/UWER/WER in \% of the different models on the context biasing text sets. $N$ is the number of distractors. Numbers in bold: Best performance for the corresponding test set, metric and number of distractors (only when is context present).}
\label{table:results}

\footnotesize

\begin{tabular}{|m{1.3cm}|c|c|c|c|c|c|c|}
\hline
Testset & \multicolumn{2}{c|}{\diagbox[width=3.0cm]{Context / N}{Approach}} & \makecell{Whisper large-v2 (1.5B)\\+ Context biasing (A)} & \makecell{Whisper large-v2 (1.5B)\\+ Context biasing (B)} & \makecell{Qwen3-ASR\\(1.7B) (C)} & \makecell{Qwen3-Omni\\(30B) (D)} & \makecell{VibeVoice-ASR\\(9B) (E)}\\
\hline
\multirow{8}{*}{\makecell{Earnings-21}} & \ding{55} & 250 & 21,99/13,03/14,25 & 28,98/13,94/15,99 & 19,99/13,10/14,04 & 26,29/17,25/18,48 & 22,62/12,84/14,18\\
 & \ding{55} & 100 & 19,99/12,85/13,82 & 25,60/13,70/15,33 & 25,32/13,56/15,17 & 28,06/18,88/20,14 & 21,76/12,76/13,99\\
 & \ding{55} & 10 & 19,07/12,70/13,57 & 19,87/12,70/13,68 & 21,59/13,36/14,48 & 30,76/20,38/21,80 & 22,85/13,70/14,95\\
 & \ding{55} & 0 & 18,73/12,69/13,52 & 18,73/12,69/13,52 & 21,59/13,39/14,51 & 30,81/21,27/22,57 & 24,57/15,46/16,71\\
\cline{2-8}
 & \ding{51} & 0 & 11.74/12.48/12.38 & 9.79/12.70/12.30 & 11.28/12.73/12.53 & 13.12/19.12/18.30 & \textbf{9.39}/\textbf{12.17}/\textbf{11.79}\\
 & \ding{51} & 10 & 11.74/12.50/12.40 & \textbf{9.79}/12.73/12.33 & 11.28/12.75/12.55 & 15.41/19.09/18.59 & 12.60/\textbf{12.08}/\textbf{12.16}\\
 & \ding{51} & 100 & 11.86/12.51/\textbf{12.42} & \textbf{10.54}/13.25/12.88 & 13.00/12.79/12.82 & 17.07/17.94/17.82 & 14.43/\textbf{12.18}/12.48\\
 & \ding{51} & 250 & 11.91/12.57/\textbf{12.48} & \textbf{11.23}/13.50/13.19 & 15.69/12.80/13.20 & 18.44/16.68/16.93 & 16.38/\textbf{12.22}/12.79\\
\hline
\multirow{8}{*}{\makecell{LibriSpeech\\test-clean}} & \ding{55} & 250 & 25,63/2,92/4,05 & 25,74/3,15/4,27 & 15,55/2,38/3,04 & 12,27/2,16/2,67 & 20,70/2,48/3,39\\
 & \ding{55} & 100 & 23,44/2,75/3,78 & 23,77/2,86/3,90 & 16,65/2,29/3,01 & 11,61/2,09/2,56 & 20,59/2,55/3,45\\
 & \ding{55} & 10 & 23,11/2,63/3,66 & 23,11/2,64/3,66 & 16,65/2,23/2,95 & 11,50/2,07/2,54 & 22,34/3,07/4,03\\
 & \ding{55} & 0 & 23,33/2,63/3,66 & 23,33/2,63/3,66 & 17,20/2,24/2,98 & 11,72/2,08/2,56 & 22,45/2,49/3,49\\
\cline{2-8}
 & \ding{51} & 0 & 7.12/2.52/2.75 & 4.60/2.55/2.66 & 6.24/2.09/2.29 & \textbf{1.10}/\textbf{1.94}/\textbf{1.90} & 2.96/2.90/2.91\\
 & \ding{51} & 10 & 7.12/2.52/2.75 & 4.60/2.56/2.67 & 5.59/2.11/2.28 & \textbf{1.75}/\textbf{1.93}/\textbf{1.92} & 6.02/2.73/2.89\\
 & \ding{51} & 100 & 7.12/2.62/2.85 & 4.71/2.75/2.85 & 8.21/2.23/2.53 & \textbf{2.30}/\textbf{2.04}/\textbf{2.05} & 7.89/2.38/2.65\\
 & \ding{51} & 250 & 7.23/2.69/2.92 & 4.71/3.03/3.11 & 10.95/3.05/3.44 & \textbf{3.83}/\textbf{2.04}/\textbf{2.13} & 9.86/3.01/3.35\\
\hline
\multirow{8}{*}{\makecell{LibriSpeech\\test-other}} & \ding{55} & 250 & 39,01/6,70/8,60 & 38,92/6,80/8,69 & 33,33/4,09/5,81 & 22,53/3,34/4,47 & 39,19/6,27/8,22\\
 & \ding{55} & 100 & 38,55/6,23/8,14 & 38,64/6,36/8,26 & 32,69/4,03/5,72 & 23,90/3,44/4,64 & 39,65/6,41/8,37\\
 & \ding{55} & 10 & 36,81/5,99/7,80 & 37,09/6,08/7,91 & 31,68/3,94/5,58 & 24,91/3,53/4,79 & 39,47/6,44/8,39\\
 & \ding{55} & 0 & 36,90/5,99/7,82 & 36,90/5,99/7,82 & 31,50/3,96/5,58 & 22,99/3,30/4,46 & 41,48/6,68/8,73\\
\cline{2-8}
 & \ding{51} & 0 & 12.55/5.66/6.07 & 10.44/5.68/5.96 & 10.26/3.56/3.95 & \textbf{3.57}/\textbf{3.33}/\textbf{3.35} & 7.88/5.66/5.79\\
 & \ding{51} & 10 & 12.45/5.67/6.07 & 10.62/5.75/6.03 & 10.44/\textbf{3.62}/4.03 & \textbf{4.76}/\textbf{3.62}/\textbf{3.69} & 16.67/5.90/6.54\\
 & \ding{51} & 100 & 12.45/5.82/6.21 & 10.62/6.12/6.39 & 14.10/3.79/4.40 & \textbf{6.32}/\textbf{3.46}/\textbf{3.63} & 21.25/5.71/6.63\\
 & \ding{51} & 250 & 12.64/5.90/6.29 & 10.71/6.48/6.73 & 21.06/3.96/4.97 & \textbf{8.42}/\textbf{3.36}/\textbf{3.66} & 24.63/5.95/7.05\\
\hline
\multirow{8}{*}{\makecell{Yodas}} & \ding{55} & 250 & 50,07/5,06/5,94 & 52,03/5,03/5,94 & 57,79/6,02/7,03 & 48,88/7,75/8,55 & 59,46/8,09/9,09\\
 & \ding{55} & 100 & 48,80/5,00/5,85 & 49,73/4,97/5,84 & 58,01/5,91/6,93 & 49,66/7,76/8,57 & 58,71/8,08/9,06\\
 & \ding{55} & 10 & 48,37/4,96/5,80 & 48,18/4,95/5,79 & 57,43/5,75/6,75 & 51,62/8,11/8,96 & 58,77/8,15/9,14\\
 & \ding{55} & 0 & 48,07/4,95/5,79 & 48,07/4,95/5,79 & 56,80/5,70/6,69 & 51,57/7,11/7,97 & 60,19/8,63/9,63\\
\cline{2-8}
 & \ding{51} & 0 & 16.90/4.98/5.22 & 5.73/\textbf{4.84}/\textbf{4.86} & 24.52/5.50/5.87 & 5.08/7.11/7.07 & \textbf{4.17}/7.74/7.67\\
 & \ding{51} & 10 & 17.06/5.00/5.23 & \textbf{5.75}/\textbf{4.84}/\textbf{4.86} & 20.78/5.60/5.89 & 9.58/7.71/7.74 & 11.66/7.85/7.92\\
 & \ding{51} & 100 & 17.19/5.04/5.27 & \textbf{6.18}/\textbf{4.87}/\textbf{4.90} & 41.27/5.86/6.55 & 20.86/7.49/7.75 & 21.37/7.86/8.12\\
 & \ding{51} & 250 & 17.23/5.07/5.31 & \textbf{6.69}/\textbf{4.93}/\textbf{4.96} & 48.32/5.99/6.81 & 26.98/7.58/7.96 & 27.92/7.89/8.28\\
\hline
\end{tabular}

\end{table*}

\subsection{Data}
\label{sec:data}

We evaluate on three test sets: Earnings-21\cite{del2021earnings}, LibriSpeech\cite{panayotov2015librispeech}, and Yodas\cite{li2023yodas}.

For Earnings-21 we extracted rare words from the given annotations: named entity (of persons), acronym (abbreviations), and domain-specific special word (products, events, laws, locations, and organizations).

For LibriSpeech we followed \cite{le2021contextualized} and used as rare words all words in the reference that fall outside 10\% of the most common words of our training data.
Then, we filtered those rare words that occurred in at least two utterances.

For our Yodas test set, we took the English data of the Yodas data set and applied a similar procedure compared to LibriSpeech: We retained words occurring at least four times but exclusively within one YouTube video (to filter noise).

This resulted in 637, 726, 821, and 6363 utterances containing at least one rare word each with a total length of 1.23, 1.91, 1.91, and 47.8 hours for Earnings-21, LibriSpeech test-clean, LibriSpeech test-other, and Yodas, respectively. In total, there are 251, 339, 353, and 1360 unique rare words, respectively.


\subsection{Metrics}
\label{sec:metric}

The performance of an ASR system is typically measured using the word error rate (WER). To measure how well a context biasing method is working, \cite{le2021contextualized} extended this metric to UWER (unbiased WER measured on words not in the biasing list) and BWER (biased WER measured on words in the biasing list), given a test set together with a corresponding context biasing list.
We evaluate these metrics along with WER to compare different approaches.

\section{Results}

The results can be seen in Table \ref{table:results}. We report BWER, UWER and WER evaluated on the test sets Earnings-21, LibriSpeech and Yodas. For each test set we report scores with and without relevant context and optional up to 250 distractors taken from the other context biasing words of the corresponding test set.

The BWER of context biasing model A improves between 37\% and 70\% relative for the different test sets when adding context for $N=0$. This is a substantial improvement. The UWER improves slightly up to 6\% relative. Adding distractors increases BWER/UWER only up to 2\%/7\% relative, in case the correct word is present in the context biasing list, and up to 17\%/12\% in the other case.

The BWER of context biasing model B is better compared to context biasing model A. It improves between 48\% and 88\% relative when adding context for $N=0$ compared to the baseline performance. The UWER also improves slightly up to 5\% relative. Adding distractors increases BWER/UWER by up to 17\%/19\% relative, in case the correct word is present in the context biasing list, and up to 55\%/20\% in the other case. Therefore, the performance of model B suffers more when adding distractors compared to model A. Most of the additional errors stem from new insertion errors. This is because the model wrongly predicts the added dynamic token.
The largest difference between model A and B is in BWER for Yodas. We believe this is because of a training-inference mismatch between the lengths of the shorter Common voice audio samples and the longer Yodas audio samples.

For the models Qwen3-ASR (C) and Qwen3-Omni (D) the baseline performance (WER, no context, $N=0$) is better on the LibriSpeech (2.98\%, 2.56\% vs. 3.66\% and 5.58\%, 4.46\% vs. 7.82\%) test sets and worse on the Earnings-21 (14.51\%, 22.57\% vs. 13.52\%) and Yodas (6.69\%, 7.97\% vs. 5.79\%) test sets compared to Whisper large-v2. The model VibeVoice-ASR (E) is only slightly better for LibriSpeech test-clean (3.49\% vs. 3.66\%) and worse for all other test sets (16.71\% vs. 13.52\%, 8.73\% vs. 7.82\% and 9.63\% vs. 5.79\%). This suggests that the speech LLMs work well for read speech but they do not generalize as well as Whisper on non-read speech.
Adding more diverse training data could help.

The BWER performance of the speech LLMs with context and without distractors improves significantly over the baseline. On the Earnings-21 and Yodas test sets, E is the best with 9.39\% and 4.17\%, respectively, and on LibriSpeech D is best with 1.1\% and 3.57\% for test-clean and test-other. However, when 250 distractors are added, the performance drops dramatically (between 39\% and 570\% relative).
Model C is robust against 10 distractors and actually performs better on LibriSpeech test-clean and Yodas compared to zero distractors. However, it has a lower score without distractors compared to models D and E. For 10 distractors and on Earnings-21 and Yodas, model B is better than all speech LLMs. When adding more distractors, the gap widens significantly.

For the models C, D and E the BWER performance without relevant context does not change significantly when adding distractors.

\begin{table*}[ht]
\normalsize
\centering

\caption{Results of context biasing list filtering for each test set: Number of utterances (Utts), true positives (TP), false positives (FP), false negatives (FN), average number of context biasing words per utterance of the ground truth (GT) and after filtering (Filt), precision and recall.}
\label{tab:results2}

\begin{tabular}{lrrrrrrrr}
\toprule
\textbf{Test set} & \textbf{Utts} & \textbf{TP} & \textbf{FP} & \textbf{FN} & \textbf{Avg GT} & \textbf{Avg Filt} & \textbf{Prec. (\%)} & \textbf{Rec.
(\%)}\\
\midrule
Earnings-21          &  637 &  813 & 1569 &  54 & 1.4 & 3.7 & 34.1 & 93.8\\
LibriSpeech test-clean &  726 &  922 &  162 &  11 & 1.3 & 1.5 & 85.1 & 98.8\\
LibriSpeech test-other &  821 & 1071 &  723 &  33 & 1.3 & 2.2 & 59.7 & 97.0\\
Yodas             & 6363 & 6184 &  417 & 390 & 1.0 & 1.0 & 93.7 & 94.1\\
\bottomrule
\end{tabular}

\end{table*}

\begin{table*}[ht]
\centering

\caption{Results: BWER/UWER/WER in \% of the different models on the context biasing text sets \emph{after} filtering the context biasing list.}
\label{tab:results3}

\footnotesize

\begin{tabular}{|m{1.3cm}|c|c|c|c|c|}
\hline
Testset & \multicolumn{2}{c|}{\diagbox[width=2.1cm]{C / N}{Approach}} & \makecell{Qwen3-ASR\\(1.7B) (C)} & \makecell{Qwen3-Omni\\(30B) (D)} & \makecell{VibeVoice-ASR\\(9B) (E)}\\
\hline
\multirow{4}{*}{\makecell{Earnings-21}} & \ding{51} & 0 & 12.49/12.82/12.77 & 15.35/19.56/18.98 & 10.77/12.45/12.22\\
 & \ding{51} & 10 & 12.54/12.78/12.75 & 15.12/19.59/18.98 & 11.00/12.46/12.26\\
 & \ding{51} & 100 & 11.63/12.78/12.63 & 14.66/19.32/18.68 & 10.94/12.16/11.99\\
 & \ding{51} & 250 & 11.97/12.92/12.79 & 14.72/19.30/18.67 & 12.08/13.08/12.95\\
\hline
\multirow{4}{*}{\makecell{LibriSpeech\\test-clean}} & \ding{51} & 0 & 6.57/2.12/2.34 & 1.86/1.91/1.91 & 3.50/2.90/2.93\\
 & \ding{51} & 10 & 6.35/2.12/2.33 & 1.53/1.91/1.89 & 3.07/2.89/2.90\\
 & \ding{51} & 100 & 6.35/2.14/2.35 & 1.31/1.90/1.87 & 3.61/2.92/2.95\\
 & \ding{51} & 250 & 6.02/2.13/2.32 & 1.31/1.96/1.92 & 3.72/2.94/2.98\\
\hline
\multirow{4}{*}{\makecell{LibriSpeech\\test-other}} & \ding{51} & 0 & 10.90/3.54/3.98 & 5.13/3.25/3.36 & 8.79/5.64/5.83\\
 & \ding{51} & 10 & 10.81/3.58/4.00 & 4.95/3.27/3.37 & 9.07/5.73/5.93\\
 & \ding{51} & 100 & 10.35/3.60/3.99 & 5.22/3.32/3.43 & 10.26/5.81/6.07\\
 & \ding{51} & 250 & 10.62/3.53/3.95 & 5.77/3.43/3.57 & 10.53/5.86/6.13\\
\hline
\multirow{4}{*}{\makecell{Yodas}} & \ding{51} & 0 & 25.92/5.51/5.91 & 8.61/7.29/7.31 & 8.17/7.80/7.81\\
 & \ding{51} & 10 & 25.73/5.51/5.91 & 8.46/7.04/7.07 & 8.08/7.80/7.81\\
 & \ding{51} & 100 & 25.45/5.49/5.88 & 8.66/7.21/7.24 & 8.30/7.79/7.80\\
 & \ding{51} & 250 & 25.72/5.52/5.92 & 8.71/7.09/7.12 & 8.71/7.81/7.82\\
\hline
\end{tabular}

\end{table*}


Since speech LLMs are sensitive to an increase in distractor count, we filtered the context biassing list. This was done by prompting Qwen3-Omni for each audio sample and each word in the context biasing list to predict if the word is present in the audio sample or not. The results can be seen in Table \ref{tab:results2}. With recall between 93.8\% and 98.8\% the model performs well in determining whether a word is present in the audio sample.
For our use case, high recall is more important than precision because only in case a relevant word is present in the prompt, it can improve BWER. On the other hand, even though the precision is not as high as recall, 
the average number of context biasing words per utterance after filtering is small (1.0-3.7).

Afterwards, we gave the filtered context biasing list to the speech LLMs. The results can be seen in Table \ref{tab:results3}. The scores without relevant context did not change significantly; therefore, we omit them. We see that with filtering there is no dramatic decrease in BWER performance because the speech LLMs receive only a few biasing words. However, in contrast to before, the BWER performance without distractors is worse than model B on Earnings-21 and Yodas (10.77\% vs. 9.79\% and 8.17\% vs. 5.73\%).

\subsection{Limitations}

The main practical drawback of the speech LLMs is that they require the context to be filtered to avoid dramatic degradation with large lists. This filtering step is computationally expensive since an additional pass over each audio–word pair is necessary, and a reliable model capable of judging word presence is needed.

At the same time, speech LLMs are considerably more flexible than the context biasing methods. Their context is not restricted to a list of words: background descriptions and other free-form prompts can all be supplied. Moreover, they can exploit semantic context rather than only surface word matches. For example, given cues such as Eastern Asia, Pacific Rim, or Indo-China, a speech LLM can favor Far East over the acoustically similar forest, a form of disambiguation the word-level biasing methods cannot perform. Our evaluation isolates the word-list setting to keep the comparison controlled and therefore does not measure these broader capabilities.

Finally, the models differ substantially in scale. While models B and C are similar in size, model B performs better overall. Even though models D and E are much larger (30B and 9B), they perform worse on Earnings-21 and Yodas.

\section{Conclusion}

We compared two Whisper-based context biasing methods (A and B) against three speech LLMs - Qwen3-ASR, Qwen3-Omni, and VibeVoice-ASR - for the recognition of rare and new words, evaluating on read (LibriSpeech) and non-read (Earnings-21, Yodas) speech in terms of BWER, UWER, and WER.
The context biasing methods reliably reduce BWER - up to 70\% relative for method A and up to 88\% for method B - while leaving UWER essentially unchanged. They remain reliable across distractor counts, since irrelevant entries are largely ignored. Method B attains lower BWER than method A when no distractors are present, but is more sensitive to their addition, with most of the additional errors arising from insertions of a wrongly predicted dynamic token.

The speech LLMs behave differently. On read speech they are strong, with Qwen3-Omni reaching the best BWER on both LibriSpeech conditions and VibeVoice-ASR the best on Earnings-21 and Yodas in the distractor-free setting. However, they generalize less well to non-read speech, where even their unbiased baselines trail Whisper, and their biasing behavior is sensitive to prompt composition: BWER degrades sharply (between 39\% and 570\% relative) as distractors accumulate, and it further depends on the order in which words appear in the prompt. When only a few distractors are present, method B already outperforms all three speech LLMs on Earnings-21 and Yodas, and the gap widens as the list grows.

Filtering the biasing list with an auxiliary model (here Qwen3-Omni) recovers much of this loss by reducing each prompt to a few likely-relevant words at high recall, but it does not close the gap: after filtering, the speech LLMs still trail method B on the non-read test sets. Overall, dedicated context biasing methods remain the more robust choice when a clean word list can be supplied and resilience to noisy or long lists matters, whereas speech LLMs offer a more flexible interface at the cost of prompt sensitivity and, in practice, an additional filtering stage.





\section{Acknowledgment}

\ifreview
BLIND
\else
The projects on which this research is based were funded by
the Horizon research and innovation program of the European Union under grant agreement No 101135798 (Meetween) and 101213369 (DVPS),
and the KIT Campus Transfer GmbH (KCT) staff in accordance with the collaboration with Carnegie – AI.
The authors gratefully acknowledge the support.
\fi

\section{Use of generative AI tools}
\ifreview
BLIND
\else
Generative AI tools were used in a limited capacity during the preparation of this work. Specifically, AI-assisted code completion was employed to support software development tasks. Language model suggestions were used to refine the clarity and style of the written text. Additionally, generative AI tools assisted in the enhancement of figures. All substantive intellectual contributions, including the research design, methodology, analysis, and conclusions, are entirely the authors' own.
\fi

\bibliographystyle{IEEEtran}
\bibliography{mybib}

\end{document}